%% file: main.tex
\documentclass{article}
\pdftrailerid{}
\PassOptionsToPackage{numbers,compress}{natbib}
\usepackage[preprint]{neurips_2020}   %

\usepackage[utf8]{inputenc}   %
\usepackage[T1]{fontenc}      %
\usepackage{hyperref}         %
\usepackage{url}              %
\usepackage{xurl}             %
\usepackage{booktabs}         %
\usepackage{mathtools,amssymb}
\usepackage{amsfonts}         %
\usepackage{nicefrac}         %
\usepackage{microtype}        %
\usepackage{array,colortbl}
\usepackage{xcolor}
\usepackage{graphicx}         %
\usepackage{algorithm,algorithmicx,algpseudocode}
\usepackage[capitalise]{cleveref}
\definecolor{saLink}{HTML}{2F5D8A}
\hypersetup{colorlinks=true,linkcolor=saLink,citecolor=saLink,urlcolor=saLink,pdftitle={Spatial Atlas: Compute-Grounded Reasoning for Spatial-Aware Research Agent Benchmarks},pdfauthor={Arun Sharma}}
\usepackage{placeins}
\usepackage{caption}
\usepackage{subcaption}
\usepackage{etoolbox}
\usepackage{enumitem}

\newcommand{\E}{\mathbb{E}}

\newcommand{\bu}{\mathbf{u}}

\newcommand{\bw}{\mathbf{w}}

\usepackage{tikz}
\usetikzlibrary{arrows.meta,positioning,shapes.geometric,shapes.misc,fit,backgrounds,calc,decorations.pathreplacing}
\usepackage{pgfplots}
\pgfplotsset{compat=1.18}
\usepackage{siunitx}
\makeatletter
\renewcommand{\@notice}{}
\makeatother
\title{Spatial Atlas: Compute-Grounded Reasoning for Spatial-Aware Research Agent Benchmarks}

\author{Arun Sharma\\University of Minnesota\\\texttt{arunshar@umn.edu}}

\begin{document}

\maketitle

\begin{abstract}
We describe compute-grounded reasoning (CGR), a design pattern in which code computes selected sub-problems from explicit intermediate representations before a language model answers. Spatial Atlas implements CGR as an Agent2Agent (A2A) server with a spatial question-answering handler and a machine-learning engineering handler. The spatial handler asks a language model to extract a scene graph, and code then fills in missing distances and checks the extracted safety rules. A separate benchmark driver can also run a strict metric bridge. It computes the gap for horizontal-gap questions from segmentation masks and a reconstructed point map, and it passes that gap to the answering model as a fact. The bridge returns a fixed unavailable answer when an evidence check fails, and it never falls back to model-estimated coordinates. The ML-engineering handler generates pipeline code and parses validation scores, and it caps the number of repair and refinement passes. Its code execution is off by default. The repository also provides four run modes that can write label-free journals, a shuffled-image control mapping, and journal validators that reject label-bearing fields. We report one private label-free operational run in which four paths each wrote eight prediction rows with zero retries. Labels stayed sealed and no score was computed, so this run establishes operational integrity only. We report no FieldWorkArena result, because the benchmark data were not accessible. We also omit every performance, latency, and resource-use number that lacks a reproducible run artifact.
\end{abstract}

\section{Introduction}
\label{sec:intro}

GPT-4 performs at a human level on various professional and academic benchmarks \citep{openai2023gpt4}. Building reliable autonomous agents on large language models (LLMs) remains an open problem \citep{wang2024survey}. Two recent benchmarks cover different parts of this problem. FieldWorkArena \citep{fieldworkarena2025} evaluates agents on field-work tasks built from images and videos captured in factories, warehouses, and retail sites. MLE-bench \citep{chan2024mlebench} tests end-to-end machine learning engineering on 75 Kaggle competitions.

This paper describes one server that hosts a spatial question-answering handler and an ML-engineering handler. The two handlers share a model client, model tiers, and usage tracking. Both kinds of task contain sub-steps that code can compute, such as a distance between two objects or a validation score.

We present \textbf{Spatial Atlas}, a research agent that exposes both handlers through a single Agent2Agent (A2A) protocol server \citep{google2024a2a}. FieldWorkArena motivated the original spatial adapter, but the benchmark is not part of the reported evaluation because its data were not accessible. The system follows a design pattern that we call \emph{compute-grounded reasoning} (CGR). When a sub-problem has a deterministic solution, code computes it first and passes the result to the language model as a fact. The architecture has the seven components below.

\begin{enumerate}[leftmargin=*,nosep]
  \item \textbf{Spatial Scene Graph Engine.} The Strong tier extracts entities and relations from the file descriptions. Code computes the distances the model did not supply, checks the extracted safety rules, and places the results in the prompt. The graph does not record whether a stored distance came from the model or from code, and perception errors can still propagate into it.
  \item \textbf{Confidence-Gated Refinement.} One Fast-tier self-grade decides whether the Strong tier writes a refinement, and the controller never refines more than once. The codebase also defines an expected-information-gain action selector, but no code in the repository calls it. The effect of this refinement on accuracy and resource use has not been measured.
  \item \textbf{Fail-Closed ML Pipeline.} This component generates strategy-aware code, and it runs that code only after an execution opt-in, an isolated-worker attestation, and authenticated server startup. The initial script and its repairs get at most 3 attempts, and score-driven refinement adds at most 2 more runs. Every run has a 600-second timeout, and dummy submissions require a separate opt-in.
  \item \textbf{Score-Driven Refinement.} An iterative loop parses machine-readable validation scores, asks the Strong tier for a revision, and keeps the revision only when its parsed score is better.
  \item \textbf{Leak Audit and Hint Registry.} A fixed audit prompt asks generated pipelines to check four common train and test leakage patterns, and a registry adds a targeted hint for its one registered competition.
  \item \textbf{Strict Metric Perception Bridge.} For horizontal-gap questions, the bridge computes the gap between two objects from segmentation masks and a reconstructed point map, and it passes that gap to the model as a fact. It records provenance and evidence digests, and it returns an explicit unavailable answer when an evidence check fails. Only the frozen benchmark driver reaches it.
  \item \textbf{Label-Free Run Modes and Validators.} The driver provides four run modes, a digest-bound shuffled-image control mapping with no fixed point, and append-only journals that reject label-bearing fields.
\end{enumerate}

The scene-graph engine and the strict metric bridge carry out compute-grounded reasoning most directly. The scene-graph engine computes distances and rule violations, and the strict metric bridge computes a reconstructed surface gap. The system writes these values into the prompt before the model answers, so an operator who records the prompt can inspect them. Their effect on reliability, accuracy, latency, and resource use must still be established through the planned evaluation.

\Cref{sec:architecture,sec:spatial,sec:entropy,sec:ml,sec:implementation} describe the system as the released code implements it. \Cref{sec:evaluation} states the evidence boundary and reports one label-free operational run, and \cref{sec:planned} gives the evaluation that remains to be done. \Cref{sec:discussion} lists the limitations.

\section{Related Work}
\label{sec:related}

\paragraph{Agent frameworks.}
AutoGPT \citep{autogpt2023} was an early open-source agent that chains LLM calls toward a user-set goal. OpenHands \citep{wangOpenHands} is an open platform for software-development agents, and it was first released under the name OpenDevin. SWE-bench \citep{jimenez2024swebench} tests whether language models can resolve real GitHub issues, and its authors report that the best model they evaluated resolved only a small share of them. SWE-agent \citep{yang2024sweagent} adds an agent-computer interface for the same task. Spatial Atlas places a spatial handler and an ML-engineering handler behind one A2A server, and the two handlers share a model client, model tiers, and usage tracking.

\paragraph{Spatial reasoning in vision-language models.}
General-purpose vision-language models (VLMs), such as models trained with visual instruction tuning \citep{liu2024llava}, answer free-form questions about images. VLMs remain weak at quantitative spatial reasoning, such as estimating distances and size differences between objects \citep{chen2024spatialvlm, liao2024qspatial, cho2026spatialclaw}. One study shows that VLMs often describe objects that are not present in the image \citep{li2023evaluating}. SpatialVLM \citep{chen2024spatialvlm} attempts to address the spatial weakness through specialized spatial training data. Our approach moves selected computations into an explicit representation, and it still depends on the accuracy of entity extraction and geometric reconstruction.

\paragraph{Scene graphs for visual reasoning.}
Visual Genome \citep{krishna2017visual} and the GQA dataset \citep{hudson2019gqa} represent visual scenes as graphs of objects and relationships. Scene graph generation \citep{xu2017scenegraph} predicts such graphs from images, and \citet{hildebrandt2020scene} answer visual questions by reasoning over scene graphs. Our scene-graph engine adapts these ideas to industrial scenes, and it adds distance computation and safety-rule checks as explicit operations.

\paragraph{Spatial representations and explicit computations.}
\citet{farhadloo2024spatiallylucid} classify multi-category point sets through spatial arrangements and training strategies that account for differences between place-types. The Atlas-EHR vision proposes a spatial hierarchy for navigating biomedical histories and identifies open research questions for decision support~\citep{farhadloo2024atlasehr}. Taxonomy-aware colocation mining computes frequent proximity relationships across a feature hierarchy~\citep{gupta2022taxonomy}, and super-colocation mining measures the density of interactions among colocated feature instances~\citep{an2025supercolocation}. These works define spatial representations and the operations that turn them into analytical outputs. Spatial Atlas applies a related separation between representation and computation when it extracts a scene graph and then computes distances or checks rules.

\paragraph{Physical bounds and incomplete spatial evidence.}
\citet{sharma2022tighterrendezvous} use time slicing to tighten bounds on possible rendezvous locations in spatial networks during trajectory gaps. \citet{sharma2024abnormalgap} combine movement bounds with a signal-coverage model to detect abnormal trajectory gaps. These methods distinguish possible movement from directly observed movement through explicit physical and spatial assumptions. Spatial Atlas addresses a different task, but its strict metric bridge also requires declared evidence before it supplies a geometric value to the answering model. Spatial Atlas does not implement these trajectory algorithms, colocation miners, spatial classifiers, or biomedical interfaces.

\paragraph{AutoML and competition-oriented systems.}
Automated machine learning frameworks such as AutoGluon \citep{erickson2020autogluon}, Auto-Sklearn \citep{feurer2019autosklearn}, and AutoKeras \citep{jin2023autokeras} aim to automate the end-to-end ML pipeline. \citet{hollmannCAAFE} use a language model to write feature-engineering code for tabular data. Our ML path adds strategy-aware code generation and bounded repair attempts behind a fail-closed execution gate.

\paragraph{A2A protocol and agent interoperability.}
The A2A protocol from Google \citep{google2024a2a} defines a standard for communication between agents, so that different agents can work together through a common interface. Our system implements an A2A server with the official \texttt{a2a-sdk}, and it exposes both spatial reasoning and ML pipeline capabilities through one task interface.

\paragraph{Information-theoretic reasoning.}
Active learning \citep{settles2009active} studies how to choose the most informative examples to label. Bayesian experimental design \citep{chaloner1995bayesian} chooses experiments that maximize an expected utility, and expected information gain is one standard choice of utility. Active prompting \citep{diaoActive} uses model uncertainty to choose which questions receive human-written chain-of-thought examples. Our design proposes an entropy-guided extension of this idea to agent action selection, which would estimate which reasoning step most reduces uncertainty about the final answer. No code in the repository calls that selector, so no result here shows its effect.

\section{System Architecture}
\label{sec:architecture}

Spatial Atlas runs as a research agent behind a dual-domain A2A server. The server receives task requests through the protocol and routes each one to a domain handler. \Cref{fig:architecture} shows the overall design.

\input{figures/fig_architecture.tex}

\paragraph{Domain classification.}
The domain classifier inspects task metadata and attachment types. The FieldWorkArena adapter recognizes its documented goal shape, and MLE-bench tasks arrive with \texttt{tar.gz} attachments that hold competition data and description files. Classification uses four fixed rules and no model call. The implementation has not been benchmarked for routing latency or resource use.

\subsection{Shared Infrastructure}
\label{sec:shared}

Both domain handlers share the components described in this subsection.

\paragraph{LiteLLM wrapper.}
We use LiteLLM \citep{litellm2024} to call models from several providers through one interface, and it records provider-reported usage when the provider supplies it. Provider reports and local estimates are not treated as exact tokenizer-equivalent counts.

\paragraph{Model tiers.}
We define four model tiers (Fast, Standard, Strong, and Vision), and each tier has a distinct role, as shown in \cref{tab:model_tiers}. Each call site names its tier in code, and no component routes calls between tiers at run time. A tier names a role, and an operator can point the Strong tier at a different provider. \Cref{sec:entropy} states which tiers the spatial path invokes and in what order.

\begin{table}[t]
\centering
\caption{This table lists the configured model tiers and their uses in the current code. It is a design table, and it reports no performance or resource-use result.}
\label{tab:model_tiers}
\begin{tabular}{@{}ll>{\raggedright\arraybackslash}p{7.4cm}@{}}
\toprule
\textbf{Tier} & \textbf{Model} & \textbf{Current use} \\
\midrule
Fast     & GPT-4.1-mini & Self-grade, and entity phrases for the generic metric engine \\
Standard & GPT-4.1      & MLE competition analysis \\
Strong   & GPT-4.1      & Scene-graph extraction, spatial answers, refinement, code generation, and code repair \\
Vision   & GPT-4.1      & Descriptions of images and video frames \\
\bottomrule
\end{tabular}
\end{table}

The default configuration maps the Standard, Strong, and Vision tiers to GPT-4.1. Operators can override \texttt{ATLAS\_STRONG\_MODEL} to test a different provider. Same-model and cross-provider refinement remain planned ablations. No accuracy or resource-use result is claimed here.

\paragraph{Public agent token reservation.}
The public A2A \texttt{Agent} uses a concurrency-safe client with a per-execution token reservation ceiling, configured by default at \num{150000} tokens. That figure is a configured cap, and this paper reports no observed token usage. Before each provider call, the client estimates the prompt size and reserves tokens under a lock, as in \cref{eq:reserve}.
\begin{align}
  m' &= \min\bigl(m,\ L - U - \mathcal{R} - \hat{p}\bigr), \qquad \text{refuse the call if } m' \le 0, \notag\\
  \mathcal{R} &\leftarrow \mathcal{R} + \hat{p} + m' \quad \text{otherwise}. \label{eq:reserve}
\end{align}
Here $L = \num{150000}$ tokens per execution, and $U$ is the committed total. The symbol $\mathcal{R}$ is the outstanding reservation, $m$ is the requested completion limit, and $\hat{p}$ is the prompt estimate. After the call, $\mathcal{R}$ drops by $\hat{p} + m'$ and $U$ rises by the same amount. The counter therefore commits each whole reservation, whatever the provider later reports.

The prompt estimate is $\max(1, \lceil \mathrm{len}/4 \rceil)$ tokens for each text part. An inline image string counts as \num{2048} tokens, and an image sent to the vision call counts as $\max\bigl(\num{1024}, \min(\num{8192}, \lceil \mathrm{bytes}/\num{1024} \rceil)\bigr)$ tokens. Concurrent calls within one A2A execution cannot oversubscribe this estimated counter. Every execution builds a fresh agent, so a new execution receives a fresh counter even for the same A2A task ID. The estimate is heuristic because provider tokenizers and image accounting vary, so the ceiling is not a hard provider-token boundary. The frozen benchmark driver uses the base client, which records usage and reserves nothing. It writes each row's usage counts into its journal artifacts.

\section{Spatial Scene Graph Engine}
\label{sec:spatial}

The spatial scene-graph engine is the central component of the spatial question-answering path and of the unexecuted FieldWorkArena adapter. It makes spatial computations explicit, but it does not remove errors introduced by perception, depth estimation, object matching, or coordinate assumptions.

\subsection{Problem Formulation}

The input is an image $I$ of an industrial scene, such as a factory, a warehouse, or a retail space, together with a natural language question $q$. The task is to produce an answer $a$, which may require counting objects, estimating distances, checking spatial containment, or verifying safety compliance. Directly prompting a VLM with $(I, q)$ is unreliable, because VLMs can describe objects that are not in the image \citep{li2023evaluating} and remain weak at estimating distances \citep{chen2024spatialvlm}.

\subsection{Scene Graph Construction}
\label{sec:scenegraph}

Our approach splits the problem into three stages, which are \emph{extraction}, \emph{structuring}, and \emph{computation}.

\paragraph{Stage 1 (extraction).}
Entity extraction has up to three steps. When the optional \texttt{torch} and \texttt{transformers} packages are installed, Florence-2~\citep{xiao2024florence2} runs first and returns object counts, keyword-based protective-equipment flags, and a caption. The repository does not declare these packages, so a default install skips this step. Next, the Vision tier (GPT-4.1 by default) writes a detailed description that lists the visible objects with approximate positions and attributes. Its prompt tells it to reuse the Florence-2 counts when they exist. Finally, the Strong tier reads the first \num{8000} characters of these descriptions and returns a JSON list of entities, relations, zones, and safety rules. Florence-2 bounding boxes are computed but never passed to a model, and detection quality on industrial imagery has not been measured here.

\paragraph{Stage 2 (structuring).}
The extracted entities form a spatial scene graph $G = (V, \mathcal{E})$. Its vertices $V$ represent entities, and its edges $\mathcal{E}$ represent spatial relations, as in \cref{eq:entity,eq:relation}.
\begin{align}
  v_i &= \texttt{SpatialEntity}(\text{id}_i, \text{label}_i, \text{pos}_i, \text{attrs}_i, \text{zone}_i), \label{eq:entity} \\
  e_{ij} &= \texttt{SpatialRelation}(\text{subj}_i, \text{pred}_{ij}, \text{obj}_j, d_{ij}). \label{eq:relation}
\end{align}
Here $\text{pos}_i \in \mathbb{R}^2$ is a position that the Strong tier estimates from the text descriptions and returns in its JSON extraction. The dictionary $\text{attrs}_i$ holds visual attributes such as color, size, and state, and $\text{zone}_i$ names the semantic zone, such as a loading dock or an aisle. The value $d_{ij}$ is the distance attached to the relation. The object of a relation can also name a zone, so an edge need not join two vertices. When the extraction supplies a distance, the graph keeps that model-estimated value. Otherwise the code computes the Euclidean distance between the two positions and rounds it to two decimals, as in \cref{eq:sgdist}.
\begin{align}
  d_{ij} &=
  \begin{cases}
    \hat{d}_{ij} & \text{if the extraction supplies a distance},\\
    \operatorname{round}_2\bigl(\lVert \text{pos}_i - \text{pos}_j \rVert_2\bigr) & \text{if both positions exist},\\
    \text{undefined} & \text{otherwise.}
  \end{cases} \label{eq:sgdist}
\end{align}

\paragraph{Stage 3 (computation).}
The scene graph also exposes the three operations below. The pipeline first fills missing distances with \texttt{compute\_all\_distances}, as in \cref{eq:sgdist}, and it then calls \texttt{check\_constraints} and \texttt{to\_fact\_sheet}. No pipeline code calls \texttt{query\_near}, and only the unit tests exercise it.

\begin{itemize}[leftmargin=*,nosep]
  \item \texttt{query\_near($v$, $r$)} returns all entities within radius $r$ of entity $v$, as in \cref{eq:near}. It recomputes each distance from the stored positions and ignores any stored relation distance.
  \item \texttt{check\_constraints()} takes no argument. It reads the safety rules that the Strong tier extracted. It flags a worker whose extracted attributes mark missing protective equipment, and it flags a person-to-hazard relation whose distance is below a threshold parsed from a rule. It returns the list of violations.
  \item \texttt{to\_fact\_sheet()} serializes the graph into a structured natural-language summary for the language model. It prints a horizontal-gap relation with six decimals and any other distance with one decimal.
\end{itemize}
\begin{align}
  N_r(v) &= \{\, u \ne v \mid \lVert \text{pos}_u - \text{pos}_v \rVert_2 \le r \,\}, \label{eq:near}\\
  \operatorname{viol}(e_{ij}) &= [\ell_i \in \mathcal{P}] \wedge [\ell_j \in \mathcal{H}] \wedge [d_{ij} < r_{\text{rule}}]. \label{eq:viol}
\end{align}
In \cref{eq:viol}, the bracket $[\cdot]$ is 1 when its condition holds and 0 otherwise. The symbol $\ell_i$ is the label of entity $i$. The person set is $\mathcal{P} = \{\text{worker}, \text{person}, \text{employee}\}$, and the hazard set is $\mathcal{H} = \{\text{forklift}, \text{machinery}, \text{crane}, \text{conveyor}, \text{vehicle}\}$. The labels are lowercased and must match a set member exactly. The threshold $r_{\text{rule}}$ is the first number that is followed, after optional spaces, by the letter m in a rule whose text contains ``distance'', ``meters'', or ``from''.

The Strong tier then receives the question, the file descriptions, and the fact sheet. The fact sheet mixes model-estimated positions, model-supplied distances, and code-computed distances, so it does not remove visual estimation from the answer.

\subsection{Strict Metric Perception Bridge}
\label{sec:bridge}

The scene graph above does repeatable arithmetic over \emph{model-estimated} two-dimensional coordinates. The arithmetic is deterministic, but its inputs are not measurements. The strict metric bridge replaces those coordinates with reconstructed geometry for one question type. It computes the gap between two objects from segmentation masks and a reconstructed metric point map. Its strict parser targets the horizontal-gap questions of Q-Spatial Bench~\citep{liao2024qspatial}. Only the frozen benchmark driver reaches the bridge, as \cref{sec:generic} explains. \Cref{fig:bridge} shows its steps. Most constants below belong to a frozen geometry contract, and the code serializes that contract and hashes it with SHA-256. The 0.30 instance-area ratio in the pair-selection step sits outside that contract.

\input{figures/fig_bridge.tex}

\paragraph{Mask alignment and erosion.}
The bridge first resizes each object mask to the $w_t \times h_t$ grid of the point map with nearest-neighbor sampling. It then erodes the resized mask $\tilde{M}$ once with a closed Euclidean disk and a zero border, as in \cref{eq:erode}.
\begin{align}
  r_e &= \Bigl\lceil \max\Bigl(\frac{w_m}{w_t},\, \frac{h_m}{h_t}\Bigr) \Bigr\rceil, \qquad
  B_{r_e} = \{(u, v) \in \mathbb{Z}^2 \mid u^2 + v^2 \le r_e^2\}, \qquad
  E = \tilde{M} \ominus B_{r_e}. \label{eq:erode}
\end{align}
Here $(w_m, h_m)$ is the original mask size, and $r_e$ is the erosion radius in point-map pixels. An empty eroded mask $E$ makes the pair invalid.

\paragraph{Overlap rule.}
Two eroded masks may still share pixels. The bridge measures their overlap with the overlap coefficient $\rho$ in \cref{eq:overlap}. A pair whose coefficient is 0.05 or more is rejected. Otherwise the shared pixels are removed from both masks.
\begin{align}
  \rho(E_a, E_b) &= \frac{\lvert E_a \cap E_b \rvert}{\min(\lvert E_a \rvert, \lvert E_b \rvert)}, \qquad
  \begin{cases}
    \text{reject the pair} & \text{if } \rho \ge 0.05,\\
    E_k \leftarrow E_k \setminus (E_a \cap E_b) \text{ for } k \in \{a, b\} & \text{otherwise.}
  \end{cases} \label{eq:overlap}
\end{align}

\paragraph{Confidence filter and voxels.}
A pixel $p$ of an eroded mask contributes only if its reconstructed point $X(p)$ has three finite coordinates and its reconstruction confidence $c(p)$ is finite and exceeds 0.3. The code treats world axes 0 and 2 of the point map as the horizontal plane and axis 1 as the vertical axis. It assumes that the reconstruction is already metric and gravity-aligned, and it does not check either property. The bridge then bins the qualified points on the horizontal $x$ and $z$ axes into \SI{5}{\milli\metre} voxels with the key $\kappa$ in \cref{eq:voxel}. Each occupied voxel keeps one representative point, chosen by the rule $\pi$ in \cref{eq:rep}.
\begin{align}
  Q(E) &= \{\, p \in E \mid X(p) \text{ finite},\ c(p) \text{ finite},\ c(p) > 0.3 \,\}, \notag\\
  \kappa(p) &= \Bigl(\Bigl\lfloor \frac{X_x(p)}{0.005} \Bigr\rfloor,\ \Bigl\lfloor \frac{X_z(p)}{0.005} \Bigr\rfloor\Bigr), \label{eq:voxel}\\
  \pi(k) &= \operatorname*{arg\,max}_{p \in Q(E),\ \kappa(p) = k} \bigl(c(p),\, -\operatorname{idx}(p)\bigr). \label{eq:rep}
\end{align}
Here $\operatorname{idx}(p)$ is the flat pixel index, and point coordinates are in meters. The arg max is lexicographic, so the highest confidence wins and a tie goes to the lowest index. SAM~3 segmentation is requested with a 0.3 threshold, and the contract records that rule as at least 0.30.

\paragraph{Voxel count limits.}
The set $K(E)$ holds the occupied voxel keys of a mask. A mask with more than \num{50000} keys keeps the \num{50000} keys with the smallest SHA-256 digests, so the retained subset does not depend on sampling order. After this cap, each mask must keep at least 32 keys, as in \cref{eq:cap}, and a mask with fewer keys makes the pair invalid.
\begin{align}
  K(E) &\leftarrow
  \begin{cases}
    \text{the \num{50000} keys with the smallest } \operatorname{SHA256}(k_x, k_z) & \text{if } \lvert K(E) \rvert > \num{50000},\\
    K(E) & \text{otherwise,}
  \end{cases} \notag\\
  \lvert K(E) \rvert &\ge 32. \label{eq:cap}
\end{align}
Here $\operatorname{SHA256}(k_x, k_z)$ is the digest of the UTF-8 string of the two signed integers joined by a comma, and digests are compared as bytes.

\paragraph{Surface gap and validity.}
The sets $R_A$ and $R_B$ hold the $x$-$z$ coordinates of the voxel representatives of objects $A$ and $B$. For each representative of one object, the bridge finds its Euclidean nearest neighbor in the other object with a \texttt{scipy} \texttt{cKDTree}. It summarizes each directed distance set by its fifth percentile with linear interpolation, and it takes the smaller of the two directed values as the gap $g$ in \cref{eq:gapfull}.
\begin{align}
  \operatorname{nn}_{A \to B}(\bu) &= \min_{\bw \in R_B} \lVert \bu - \bw \rVert_2 \quad (\bu \in R_A), \notag\\
  g &= \min\Bigl(\operatorname{pct}_{5}\bigl(\{\operatorname{nn}_{A \to B}(\bu)\}_{\bu \in R_A}\bigr),\, \operatorname{pct}_{5}\bigl(\{\operatorname{nn}_{B \to A}(\bw)\}_{\bw \in R_B}\bigr)\Bigr), \notag\\
  0 &< g \le 20\ \text{m}. \label{eq:gapfull}
\end{align}
Here $\operatorname{pct}_{\alpha}$ is the linear-interpolation percentile at the zero-based position $(\alpha/100)(n-1)$ among $n$ values sorted in ascending order. A gap that is not finite or lies outside the interval $(0, 20]$~m makes the pair invalid.

The fifth percentile limits the influence of a few extreme distances while keeping the near-contact region that the question asks about. At the 32-voxel minimum it falls between the second and third smallest distances, so the single smallest distance in each direction does not set the gap. A stray reconstructed point close to the other object can still lower the gap, and its effect on gap error has not been evaluated. Taking the symmetric minimum removes the dependence on which object is named first.

\paragraph{Instance filter and pair selection.}
A distinct-pair question names two different objects, and every cross pair of their instances is a candidate. A repeated-instance question asks about two instances of the same object. For such a question, an instance enters pairing only if its mask area is at least 0.30 of the largest instance area, and at least two instances must remain. Every pair of the remaining instances is a candidate. The bridge selects the valid pair with the smallest gap and breaks ties by the lowest mask indices, as in \cref{eq:select}.
\begin{align}
  \mathcal{I} &= \{\, i \mid \max_j \operatorname{area}_j > 0,\ \operatorname{area}_i \ge 0.30 \max_j \operatorname{area}_j \,\}, \qquad
  (i^*, j^*) = \operatorname*{arg\,min}_{(i, j)\ \text{valid}} \bigl(g_{ij},\, (i, j)\bigr). \label{eq:select}
\end{align}
Here $\operatorname{area}_i$ is the pixel count of the raw segmentation mask of instance $i$, and $\mathcal{I}$ applies only to repeated-instance questions. The arg min is lexicographic. The 0.30 instance-area ratio was chosen by inspecting development cases, and its effect on gap error has not been evaluated. Because the ratio sits outside the hashed contract, a change to it would leave the contract digest unchanged.

\paragraph{Output and failure behavior.}
\Cref{eq:strictout} gives the outcome $B$ of the strict bridge.
\begin{align}
  B &=
  \begin{cases}
    g_{i^* j^*} \text{ in meters, in the fact sheet} & \text{if a valid pair exists},\\
    \texttt{QSPATIAL\_UNAVAILABLE} & \text{if an evidence check fails},\\
    \text{an error} & \text{if parsing, the manifest check, or a service fails.}
  \end{cases} \label{eq:strictout}
\end{align}
The evidence checks are an unsupported question, an empty segmentation, fewer than two plausible instances, and the absence of any valid pair. The fact sheet prints $g$ to six decimals. The constant \texttt{QSPATIAL\_UNAVAILABLE} holds the literal answer \texttt{measurement unavailable}, and it becomes the final answer. After a valid measurement, the handler passes the fact sheet to the controller in \cref{sec:controller}. The Strong tier then writes the final answer and may refine it once, so the gap reaches the answer only through the model. No code checks that the final answer repeats $g$.

Two properties of this path matter for interpretation. First, it is deliberately \emph{fail-closed}. It records protocol identifiers, provenance, evidence digests, and range checks. When an evidence check fails, it returns the literal answer \texttt{measurement unavailable} before the reasoner runs. It raises an error when the question falls outside the parser grammar, when the sample metadata fail their manifest check, or when the segmentation or reconstruction service fails. It never substitutes the estimated-coordinate scene graph in either case. Second, reconstructed geometry is an estimate and not ground truth. Segmentation errors, reconstruction scale errors, and referent mismatches all remain possible. The fail-closed checks catch only the failures that break an evidence check. A wrong mask or a wrong scale that passes every check still produces a wrong gap.

The strict parser was written against Q-Spatial Bench question text. It carries hand-written aliases for specific phrasings, it marks three rows as unsupported, and a separate table rewrites some grounding queries sent to SAM~3. That table sits outside the hashed contract. The parser is therefore fitted to this benchmark and is not a held-out component.

The bridge code is public, but it is not a stand-alone reproduction. The live perception path imports its segmentation and reconstruction tools from an externally installed copy of SpatialClaw~\citep{cho2026spatialclaw}. SpatialClaw is a separate training-free framework that uses code as its action interface in a persistent kernel. Its GPU service runs SAM~3 segmentation~\citep{carion2025sam3} and Depth Anything~3 reconstruction~\citep{lin2025da3}. Neither SpatialClaw nor its GPU service is vendored here, and the gated benchmark images and the frozen control artifact are not bundled with the repository.

\subsection{Generic Metric Engine}
\label{sec:generic}

The scene-graph engine is the default, and an operator selects the metric engine through configuration. On the A2A path, metric-engine requests use a generic metric engine. The A2A server never passes the strict protocol identifier, so an A2A request in metric mode always uses the generic engine. Only the frozen benchmark driver reaches the strict bridge. The driver can also pass exact benchmark regions, and that path uses SpatialClaw's centroid rule without a Fast-tier call. In the generic engine, the Fast tier lists up to six entity phrases to segment. For each object $i$, the engine keeps the mask points $Q_i$ whose coordinates are finite and whose confidence exceeds 0.3. It then computes a position and a visible height, as in \cref{eq:generic}.
\begin{align}
  \text{pos}_i &= \bigl(\operatorname{med}_{p \in Q_i} X_x(p),\ \operatorname{med}_{p \in Q_i} X_z(p)\bigr), \notag\\
  h_i &= \operatorname{pct}_{97.5}\bigl(\{X_y(p)\}_{p \in Q_i}\bigr) - \operatorname{pct}_{2.5}\bigl(\{X_y(p)\}_{p \in Q_i}\bigr), \qquad \lvert Q_i \rvert \ge 32,\ \ h_i > 0. \label{eq:generic}
\end{align}
The centroid is the median of the confident reconstructed points, and the engine stores its $x$ and $z$ coordinates as the entity position in the scene graph. Visible height is the 97.5th percentile minus the 2.5th percentile of the vertical coordinate. A height needs at least 32 qualified points, and a height of zero or less is rejected.

The two metric paths fail in different ways. When strict mode is off, which is the A2A default, the generic engine falls back to the scene-graph path after a failure in the metric backend. In that mode, a failed instance is skipped, and a metric scene with no entities is kept rather than replaced. The strict path never falls back, and it fails closed as described in \cref{sec:bridge}. The frozen benchmark driver turns on strict mode whenever the metric engine is selected.

\subsection{Label-Free Run Modes and Validators}
\label{sec:runmodes}

The repository provides the run modes and validators for a four-path comparison. Two options select a path. The engine option is the scene-graph engine or the metric engine, and the image-mode option is the correct image, the question only, or a shuffled image. The options are not freely combinable, because the driver enforces four constraints. The question-only mode requires the scene-graph engine, because it withholds the image and so has no geometry to measure. The shuffled-image mode requires the metric engine and an explicit control mapping, and a control mapping is rejected for any other image mode. The driver therefore rejects the pairs (metric, question-only) and (scenegraph, shuffled). Only four of the six combinations are admitted, as \cref{eq:arms} states.
\begin{align}
  \mathcal{M} &= \bigl\{(\text{scenegraph}, \text{correct}),\ (\text{scenegraph}, \text{question-only}), \notag\\
  &\qquad (\text{metric}, \text{correct}),\ (\text{metric}, \text{shuffled})\bigr\}. \label{eq:arms}
\end{align}
A control mapping is required exactly when the image mode is shuffled. The four admitted paths are the scene graph with the correct image, the question-only baseline, the metric engine with the correct image, and the metric engine with a shuffled image.

The shuffled-image control is bound to a contract. The control mapping $\phi$ is defined over the set $S$ of source image paths, and $\delta_s$ is the recorded SHA-256 digest of the image stored at path $s$. The loader checks the size of the mapping, its bijectivity, the absence of fixed points, and the digest of every source and control image, as \cref{eq:mapping} states.
\begin{align}
  &\phi : S \to S, \qquad \lvert S \rvert = 84, \qquad \phi \text{ is a bijection}, \qquad \phi(s) \ne s \ \ \forall s \in S, \notag\\
  &\operatorname{SHA256}(s) = \delta_s \ \text{ and } \ \operatorname{SHA256}\bigl(\phi(s)\bigr) = \delta_{\phi(s)} \ \ \forall s \in S. \label{eq:mapping}
\end{align}
Here $\operatorname{SHA256}(s)$ is the digest of the image stored at path $s$. The loader rejects the run if any image has drifted or if a source repeats. The mapping therefore holds exactly 84 pairs, and no source path maps to itself. The loader compares paths and does not require $\delta_s \ne \delta_{\phi(s)}$, so it would admit two byte-identical images stored under different paths. The mapping metadata must also name a fixed cyclic-next algorithm over byte-sorted paths.

Prediction rows are appended to journals under the label-free schema \texttt{label\_free\_prescore\_v1} with an append-only file flag. The writer rejects label-bearing keys anywhere in a payload, and it uses an exact-key list plus token and prefix rules to find them. Prediction metadata must contain exactly six safe fields. The writer also refuses to mix label-free and ordinary rows in one journal, and it refuses to resume a journal whose schema does not match the current run.

These validators keep label-bearing fields out of the prescore journals and the recorded metric evidence. They check what a run writes, so the claim that labels stayed sealed in the reported run rests on the run record in \cref{sec:evaluation}. The public repository does not provide the combined single-command orchestrator that runs all four paths under one root.

\paragraph{Output formatting against benchmark scorers.}
FieldWorkArena scoring is done by the benchmark's own evaluation program \citep{fieldworkarena2025}, and this repository implements none of its scoring functions. It does implement an output formatter that shapes each answer, so that a correct answer is not rejected on presentation. A right answer in the wrong shape can score zero under exact and substring matchers. The public FieldWorkArena evaluator uses string rules for its exact, include, and exclude matchers, and it asks a language model to grade its fuzzy, JSON, and numerical matchers \citep{fieldworkarena2025}.\footnote{The evaluator code is in the public repository at \url{https://github.com/FujitsuResearch/FieldWorkArena}.} Because the gated benchmark data were never accessible, this formatter has never been exercised against an official FieldWorkArena evaluation run.

\section{Confidence-Gated Refinement and a Proposed Entropy-Guided Design}
\label{sec:entropy}

This section separates the controller that the spatial path executes from an information-theoretic design that no code executes. The design draws on active learning \citep{settles2009active} and Bayesian experimental design \citep{chaloner1995bayesian}. The difference between the two parts matters for how the design should be judged.

\subsection{Implemented Controller}
\label{sec:controller}

The implemented controller answers once at the Strong tier, asks the Fast tier for one self-grade, and refines at most once, as in \cref{eq:ctrl1,eq:ctrl2,eq:ctrl3}.
\begin{align}
  a_0 &= \operatorname{Strong}\bigl(q,\, D_{:12000},\, F\bigr), \label{eq:ctrl1}\\
  \sigma &=
  \begin{cases}
    \operatorname{parse}\bigl(\operatorname{Fast}(q,\, D_{:2000},\, a_0)\bigr) & \text{if the reply parses},\\
    0.5 & \text{otherwise},
  \end{cases} \label{eq:ctrl2}\\
  a &=
  \begin{cases}
    \operatorname{Strong}\bigl(q,\, a_0,\, D_{:6000},\, F\bigr) & \text{if } n_{\text{refl}} > 0 \text{ and } \sigma < 0.6,\\
    a_0 & \text{otherwise.}
  \end{cases} \label{eq:ctrl3}
\end{align}
Here $D_{:n}$ denotes the first $n$ characters of the file descriptions, and $F$ is the fact sheet. Both Strong-tier prompts also carry the expected output format, and $F$ is empty when the scene has no entities. The value $n_{\text{refl}}$ is the reflection setting, whose default is 2 and which acts only as an on-off flag. The self-grade runs only when $n_{\text{refl}} > 0$, and \cref{alg:controller} restates the same procedure.

\begin{algorithm}[!htb]
\caption{This procedure is the implemented spatial reasoning controller, with $D_{:n}$ as defined above. The self-grade $\sigma$ is a self-reported heuristic. The code does not clip it, and it is not calibrated.}
\label{alg:controller}
\begin{algorithmic}[1]
\Require Question $q$, file descriptions $D$, fact sheet $F$, threshold $\tau = 0.6$
\State $a \gets \textsc{Strong}(q, D_{:12000}, F)$
\If{reflection is disabled}
  \State \Return $a$
\EndIf
\State $\sigma \gets \textsc{Fast-SelfGrade}(q, D_{:2000}, a)$, or $\sigma \gets 0.5$ when the reply cannot be parsed
\If{$\sigma < \tau$}
  \State $a \gets \textsc{Strong-Refine}(q, a, D_{:6000}, F)$ \Comment{runs at most once}
\EndIf
\State \Return $a$
\end{algorithmic}
\end{algorithm}

The implemented controller asks the Fast tier for one self-reported score $\sigma$ for the single Strong-tier answer. It does not approximate the information-gain rule that \cref{sec:proposed} proposes. The prompt requests a value between 0.0 and 1.0, but the code does not clip the parsed value, and a missing or unparseable score defaults to 0.5. This score has not been shown to be calibrated, so it should not be read as a posterior probability.

During reflection, the Strong tier receives the question, its first answer, the first \num{6000} characters of the file descriptions, and the fact sheet, and it writes one revised answer. The first answer had already seen the first \num{12000} characters of the same descriptions, so reflection adds no new evidence. Because the 0.5 default lies below $\tau = 0.6$, an unparseable self-grade always triggers this refinement. The configuration field named for reflection rounds acts as an enable flag, so the controller performs at most one refinement per task and never loops.

\FloatBarrier
\subsection{Proposed Entropy-Guided Design}
\label{sec:proposed}

The formulation in this subsection is a proposed design. No code in the repository computes \cref{eq:entropy} or runs the selection rule in \cref{eq:ig}.

At each reasoning step $t$, an agent under this design would keep a knowledge state $\mathcal{K}_t$ of accumulated observations, computed facts, and intermediate conclusions. The \emph{answer entropy} in \cref{eq:entropy} measures the uncertainty over the space of possible answers.
\begin{align}
  H(\mathcal{A} \mid \mathcal{K}_t) &= -\sum_{y \in \mathcal{A}} P(y \mid \mathcal{K}_t) \log P(y \mid \mathcal{K}_t). \label{eq:entropy}
\end{align}
Here $\mathcal{A}$ is the set of candidate answers, and $P(y \mid \mathcal{K}_t)$ is the estimated probability of answer $y$ given the current knowledge. The design would consider $n_c$ candidate actions $\{c_1, \ldots, c_{n_c}\}$, such as examining one region of the image, querying the scene graph, or calling a stronger model. It would then select the action with the largest expected information gain, as in \cref{eq:ig}.
\begin{align}
  c^* &= \arg\max_{c_j} \ \E_{o \sim P(o \mid \mathcal{K}_t, c_j)}\bigl[ H(\mathcal{A} \mid \mathcal{K}_t) - H(\mathcal{A} \mid \mathcal{K}_t \cup \{o\}) \bigr]. \label{eq:ig}
\end{align}
Here $o$ is the observation that action $c_j$ would return. The repository defines a selector for this rule, and that selector would ask the Fast tier to rate each candidate action from 1 to 10. No code in the repository calls it, so this paper makes no claim that the selector runs or reduces model calls.

\subsection{Model Routing}
\label{sec:routing}

A graded routing policy is the design target for efficient use of model calls. Under it, a fast tier would answer confidently handled questions, a standard tier would handle moderate cases, and a strong tier would take the hardest ones. We state this policy as a target, and it does not describe the current spatial path.

The repository also defines a tier-router class, and no code in the repository calls it. The implemented spatial controller answers directly at the Strong tier over the evidence already built, uses the Fast tier only for the self-grade, and escalates no further than one Strong-tier refinement. It does not perform graded escalation. So this paper makes no efficiency claim for the spatial path. Whether graded routing reduces model usage without degrading answers is an open question for the planned evaluation, and answering it requires the protected scoring that has not yet been performed. The effect of the routing policy on resource use is likewise unmeasured.

\section{Fail-Closed ML Pipeline}
\label{sec:ml}

The ML-engineering handler generates one pipeline script from each competition description. Execution is fail-closed, and it requires both \path{ATLAS_ENABLE_MLEBENCH_CODE_EXECUTION=true} and \path{ATLAS_TRUSTED_ISOLATED_WORKER=true}. The second flag is an operator's attestation that the process runs in an isolated, trusted worker, and the code cannot confirm that isolation. Server startup then requires an \path{ATLAS_BEARER_TOKEN} with at least 32 characters.

\paragraph{Competition analysis.}
When a competition task arrives, the Standard tier reads the description, a file listing, and a data preview. It returns JSON with a task type, a metric and its direction, a target column, the submission format, a short data summary, a strategy name, and key insights. The strategy name selects a template from \cref{tab:strategies}.

\begin{table}[t]
\centering
\caption{This table lists the strategy templates that the handler inserts into the code-generation prompt. The Strong tier writes the final script, so a template guides the code and is not the executed code.}
\label{tab:strategies}
\begin{tabular}{@{}l>{\raggedright\arraybackslash}p{3.2cm}>{\raggedright\arraybackslash}p{7.0cm}@{}}
\toprule
\textbf{Strategy name} & \textbf{Task type} & \textbf{Template contents} \\
\midrule
\texttt{tabular}    & Tabular classification or regression & AutoGluon \texttt{TabularPredictor} with a 300-second limit, which falls back to LightGBM when AutoGluon is not installed \\
\texttt{nlp}        & Text classification & TF-IDF unigram and bigram features (up to \num{50000}) with logistic regression and a cross-validation check \\
\texttt{vision}     & Image classification & Flattened $32 \times 32$ pixel features with a tree-ensemble classifier \\
\texttt{timeseries} & Forecasting & Lag features (1, 7, 14, and 28 steps), rolling features, and a LightGBM regressor \\
\texttt{general}    & Mixed or unknown & A gradient-boosting classifier or regressor \\
\bottomrule
\end{tabular}
\end{table}

\subsection{Code Generation and Execution}

For each competition, the pipeline asks the Strong tier for a single standalone Python script with the steps below.
\begin{enumerate}[leftmargin=*,nosep]
  \item The script loads and preprocesses the training data according to the detected task type.
  \item It implements the selected strategy with suitable hyperparameters.
  \item It holds out a simple validation split, trains the model, and prints one \texttt{VALIDATION\_SCORE} line for that split.
  \item It generates predictions on the test set in the required submission format.
  \item It writes a valid \texttt{submission.csv} to the expected output location.
\end{enumerate}

After explicit authorization, the generated script runs in a bounded subprocess with a 600-second timeout. The subprocess captures bounded stdout and stderr and uses a minimal environment. It is terminated as a process group on timeout or when the handler cancels the run internally. The A2A cancel operation itself is not supported. These controls are defense in depth, and they do not form a complete security sandbox.

\paragraph{Bounded repair loop.}
When an explicitly authorized execution fails, the handler runs the repair steps below.
\begin{enumerate}[leftmargin=*,nosep]
  \item \textbf{Error capture.} The executor keeps the tail of stderr and stdout from the failed run.
  \item \textbf{Repair.} The Strong tier receives the script, the last \num{2000} characters of the error, the last \num{1000} characters of stdout, and the competition description, and it returns a complete revised script. Its prompt asks it to change only what the error requires.
  \item \textbf{Re-execution.} The handler runs the revised script under the same timeout.
\end{enumerate}
The setting \texttt{max\_code\_iterations\,=\,3} allows 3 total attempts, including the initial attempt. If all attempts fail, the task fails by default. A schema-shaped dummy submission is produced only when the operator separately sets \texttt{ATLAS\_ALLOW\_DUMMY\_SUBMISSION=true}.

\paragraph{Score-driven refinement loop.}
Error recovery can repair a pipeline that crashes, but it cannot raise the score of a working pipeline. After the first authorized run succeeds, the handler parses the last match of the pattern \texttt{VALIDATION\_SCORE: <float>} in the script's standard output, and it skips refinement when the first run prints no score. Otherwise it requests one targeted revision, runs it under the same authorization and controls, and keeps it only when the parsed score is better under the metric direction. \Cref{eq:mle} states the attempt limits and this acceptance rule, with $\nu$ as the best parsed score so far and $\nu'$ as the revised score.
\begin{align}
  n_{\text{attempt}} &\le 3, \qquad n_{\text{refine}} \le 2, \qquad t_{\text{run}} \le 600\ \text{s}, \notag\\
  \text{accept } \nu' &\iff
  \begin{cases}
    \nu' < \nu & \text{if the direction text contains a minimize keyword},\\
    \nu' > \nu & \text{otherwise.}
  \end{cases} \label{eq:mle}
\end{align}
Here $n_{\text{attempt}}$ counts the initial run and its repairs, $n_{\text{refine}}$ counts the refinement runs, and $t_{\text{run}}$ is the timeout of each run. The minimize keywords are \texttt{min}, \texttt{lower}, \texttt{loss}, \texttt{error}, \texttt{rmse}, \texttt{mae}, and \texttt{mse}. The loop runs up to \texttt{max\_refinement\_iterations\,=\,2} extra passes, and it starts no new pass once \texttt{refinement\_wall\_time\_seconds\,=\,900} have elapsed since the first successful run. The 900-second check runs before each pass, so a pass that starts before that point can finish after it. The selection logic discards revisions that regress or fail to print a score.

This loop uses the configured Strong tier. The Strong tier defaults to GPT-4.1, and the operator can override it. Whether a cross-provider override outperforms a same-model retry is a planned ablation and not an established result.

\subsection{Leak Audit and Leak Hint Registry}

A competition can leak test information through training-set overlap, public dataset ancestry, or file metadata. Spatial Atlas keeps a \emph{leak hint registry} whose entries are text instructions injected into the Strong-tier code-generation prompt when a competition is detected. It does not hand-code exploit solvers, because their fixed merge keys may not match the MLE-bench tar layout.

Every code-generation call also receives a universal \emph{leak audit preamble}. The preamble asks the Strong model to run four checks before it trains any model.
\begin{enumerate}[leftmargin=*,nosep]
  \item The model compares ID-like columns between train and test for row-level overlap.
  \item It computes row fingerprints, which hash the non-target features, to detect duplicated content.
  \item It checks temporal ordering in timestamp-based competitions, where temporal shuffling can leak test information.
  \item It hashes file bytes in media-based competitions to detect identical train and test files.
\end{enumerate}
The audit text also tells the model to use the train labels for the matching rows when more than 50\% of rows overlap. The audit runs whether or not a registered entry matches. Its detection rate has not been measured, so this paper makes no claim that it finds any leak.

The registry holds one entry, for the Random Acts of Pizza competition. When its detection predicate matches, a targeted hint is appended after the generic audit text. The Strong model writes the final pandas operations against the actual tar layout it sees at run time, and the audit policy stays in a single file (\texttt{mlebench/strategies/leaks.py}).

\paragraph{Strategy selection.}
The Standard-tier analyzer picks one strategy template from the task description. The ML path does not call the entropy module (\texttt{src/entropy/engine.py}), and it does not generate candidate solutions for several strategies. Entropy-guided strategy selection is not implemented.

\section{Implementation Details}
\label{sec:implementation}

\paragraph{A2A protocol.}
Spatial Atlas implements the A2A protocol with the official \texttt{a2a-sdk}. The project requires version 0.3.20 or later, and its lock file pins version 0.3.25, which implements A2A specification v0.3.0 \citep{google2024a2a}. The server exposes an A2A endpoint that accepts JSON-RPC task submissions and returns structured results in the protocol format. It sends intermediate status updates through the SDK's task updater, and the agent card advertises streaming. The agent card also describes both FieldWorkArena and MLE-bench task types.

\paragraph{Deployment.}
The system is packaged as a Docker image built from the repository's Dockerfile. Environment variables configure API keys, model endpoints, and resource limits. Public HTTP request bodies default to a 64~MiB limit. At most four non-read-only requests run at once by default, and the server answers HTTP 503 beyond that limit instead of queueing without bound. A configured \path{ATLAS_BEARER_TOKEN} must contain at least 32 characters, and it authenticates non-read-only requests. Normal non-loopback startup requires the token even when MLE execution is disabled, and loopback development may omit it. The setting \path{ATLAS_ALLOW_UNAUTHENTICATED_PUBLIC=true} is a test-only override for non-loopback binding. MLE execution always requires authenticated startup.

\paragraph{File processing pipeline.}
Task inputs arrive in several formats, and each format has its own processing path.
\begin{itemize}[leftmargin=*,nosep]
  \item \textbf{Images.} When Florence-2 is available, it runs first and supplies object counts and a caption. The Vision tier (GPT-4.1 by default) then describes the scene. Images in RGBA, LA, or palette mode are converted to RGB, and every image is re-encoded as JPEG before the call. No resizing is applied.
  \item \textbf{PDFs.} Text is extracted page by page with \texttt{pypdf}. The pipeline has no OCR step, so a scanned page without a text layer yields no text.
  \item \textbf{Videos.} OpenCV samples one frame every two seconds, and a long video is sampled more sparsely so that at most 30 frames are kept. At most 10 evenly spaced frames are then described. No scene-change keyframe detection is performed.
  \item \textbf{Archives.} The handler extracts \texttt{tar.gz} files, which carry MLE-bench data, to a temporary workspace directory.
  \item \textbf{Text.} Files are decoded as UTF-8, and Latin-1 is used as a fixed fallback when UTF-8 decoding fails.
\end{itemize}

\paragraph{Model configuration.}
All model calls use the tiers in \cref{tab:model_tiers}. The Fast tier defaults to \texttt{openai/gpt-4.1-mini}. It produces the self-grade, and in the generic metric engine it also lists the entity phrases to segment. The Strong tier extracts the scene graph from the file descriptions, and the Vision tier (\texttt{openai/gpt-4.1}) describes images and video frames. Domain classification uses fixed rules and no model call. The Standard and Strong tiers both default to \texttt{openai/gpt-4.1}, and the Strong tier can be overridden with \texttt{ATLAS\_STRONG\_MODEL}. Any cross-provider variant must be compared with the default under identical token and time limits.

\paragraph{Resource controls.}
Public A2A model calls use the concurrency-safe per-execution reservation in \cref{eq:reserve}. The frozen benchmark driver uses the base client, which records usage and reserves nothing. Reflection performs at most one refinement per task. Generated-code execution is disabled by default, and it requires both execution flags plus authenticated server startup. After authorization, each pipeline attempt has a 600-second timeout, and the initial run plus the repair loop permits at most 3 total attempts. Dummy submissions remain disabled unless separately enabled. After a successful real run, the score-driven refinement loop may run up to 2 more passes (\cref{eq:mle}), and it starts no new pass after 900 seconds. Every number in this paragraph is a configured ceiling that bounds worst-case work. None is an observed runtime, an observed attempt count, or a measured resource use.

\section{Evaluation}
\label{sec:evaluation}

\subsection{Current Evidence Boundary}
\label{sec:boundary}

This version reports no claim-bearing benchmark table. FieldWorkArena remained gated and inaccessible, so the project did not run its validation set and reports no FieldWorkArena accuracy, ablation, latency, token, or resource-use result. The local adapter and tests establish software behavior only, and they are not benchmark evidence.

The repository also does not contain a sealed, end-to-end MLE-bench result artifact covering the full competition suite. This version removes the previously drafted valid-submission, medal, refinement, leak-effectiveness, and resource-use figures. A completed job or a working code path is not treated as a scientific result without the corresponding immutable predictions, scorer output, run manifest, and logs.

The evidence in this paper has two parts, and neither part measures answer quality. The first part is one private label-free operational run, called V37, in \cref{sec:v37}. The second part is a pair of software test results in \cref{sec:tests}.

\subsection{Label-Free Operational Run (V37)}
\label{sec:v37}

We report one completed label-free operational run of the four-path design. It is an integrity check on the execution path, and it does not measure answer quality. \Cref{fig:arms} summarizes the four arms of the run.

\input{figures/fig_arms.tex}

Four paths ran together. They were the scene-graph path, the question-only baseline, the correct-image metric path, and the schema-matched shuffled-image metric control. Each path wrote eight label-free prediction rows, for 32 rows in total. The run used one submission with zero retries and no rollback. Every bounded execution step exited cleanly, and all eight bounded service logs passed the required warning scan. Registry restoration passed, and the job-local authentication artifact was absent after teardown. The terminal finalizer ran exactly once. It created the terminal result and the smoke seal, which is the closed final record of this smoke-test run.

The shuffled arm is the record's schema-matched shuffled-image metric control. In the public driver, a shuffled-image run is admitted only under the control-mapping contract in \cref{eq:mapping}. The V37 record does not restate the mapping that the private run loaded, so this paper does not report its size. In the public driver's shuffled mode, the image files sent to the model come from the mapped control images. Whether the strict geometry step in the V37 shuffled arm measured the control image or the source image depends on the private benchmark loader. The public code does not settle that question, so we treat the shuffled arm as an executed control path. We do not treat it as a geometry control.

\begin{table}[t]
\centering
\caption{This table quotes the V37 evidence record, which asks writers to use this wording and no stronger interpretation. In item 9 on the right, this paper writes ``resource-use'' in place of the record's term.}
\label{tab:v37}
\small
\begin{minipage}[t]{0.48\linewidth}
\raggedright
\textbf{V37 establishes}\par\smallskip\hrule\smallskip
\begin{enumerate}[leftmargin=*,nosep]
  \item Four operational paths ran together.
  \item The paths were the scene-graph path, question-only baseline, correct-image metric path, and schema-matched shuffled-image metric control.
  \item Each path produced eight label-free prescore prediction rows.
  \item The total was 32 rows.
  \item The execution, warning-scan, restoration, cleanup, and terminal-seal gates passed.
  \item The run used zero retries.
  \item Protected labels and ground truth remained sealed.
  \item No score or paired comparison was computed.
  \item This establishes operational integrity only.
\end{enumerate}
\end{minipage}\hfill
\begin{minipage}[t]{0.48\linewidth}
\raggedright
\textbf{V37 does not establish}\par\smallskip\hrule\smallskip
\begin{enumerate}[leftmargin=*,nosep]
  \item Accuracy.
  \item Superiority over another path or system.
  \item A causal or ablation effect.
  \item Statistical significance.
  \item An uncertainty interval.
  \item Calibration.
  \item Benchmark ranking.
  \item Production readiness.
  \item A resource-use or latency result.
  \item Native SpatialClaw persistent-kernel integration.
\end{enumerate}
\end{minipage}
\end{table}

\Cref{tab:v37} lists what V37 establishes and what it does not. V37 establishes operational integrity only. Protected labels and ground truth were never opened, and no benchmark score, paired comparison, uncertainty interval, resource-use result, or latency result was computed. The shuffled-image path is a control-path \emph{execution}. Because the protected labels stayed sealed, its score and its paired effect against the correct-image path remain unknown.

Three distinctions matter for reading this result. First, the prediction journals record model outputs before scoring. They are not answer-level evidence-use journals, and they do not show which evidence a given answer used. Second, the terminal finalizer checks lifecycle and evidence-integrity conditions. It is not an answer verifier, and it does not check answer correctness. Third, the metric arms used SpatialClaw perception primitives through the bridge in \cref{sec:bridge}, and V37 did not run a native SpatialClaw persistent-kernel arm.

The public repository supplies the metric bridge, the four run modes, the control-mapping contract, and the label-free journal validators. The combined four-path orchestration that produced V37 ran in a separate private execution environment, and it is not part of the public repository. Every arm also needs a SpatialClaw checkout and the benchmark data, so the public repository cannot reproduce V37.

\subsection{Software Test Results}
\label{sec:tests}

We report two software test results, and neither one measures answer quality. First, the frozen V37 control suite passed 350 tests plus 63 subtests. The evidence record treats this count as a local regression pass, and it is not benchmark accuracy. Second, a local run of the public CI test command with Python 3.13 reported 323 passed and 1 deselected. The deselected case is the end-to-end test, which needs a GPU tool server and the dataset.

The same CI command reports 92.09\% line coverage of \texttt{src/fieldwork/perception.py}, and this figure covers that one module only. CI fails when that coverage drops below 90\%. The two test counts come from different suites, so they should not be compared with each other. Passing tests establish only the software behavior they exercise, and they do not promote proposed behavior into a result.

\section{Planned Evaluation Protocol (Future Work)}
\label{sec:planned}

The evaluation in this section is planned future work, and none of it has been performed. The spatial evaluation will compare a question-only baseline, the scene-graph path, the metric-perception path, and a native reference implementation on a frozen public slice. It will report per-question-type accuracy, paired uncertainty intervals, parser and geometry failure rates, model and data revisions, token usage, wall-clock latency, and artifact paths. Labels will stay sealed until all prediction journals are complete. Protected scoring, paired analysis, and uncertainty analysis are the next scientific phase.

The protected scoring protocol is specified in advance as the ordered steps below.
\begin{enumerate}[leftmargin=*,nosep]
  \item The analysis plan and the interpretation boundary will be frozen before anything is unsealed.
  \item Every prediction journal will be checked for completeness and schema validity.
  \item The protected labels will be unsealed only after that check passes.
  \item Each path will be scored with the benchmark's own metric.
  \item The correct-image metric path will be compared with the scene-graph path and with the shuffled-image control, and each paired comparison will carry an uncertainty interval.
  \item Parser failures, geometry failures, refusals, and invalid cases will be reported alongside every aggregate.
\end{enumerate}

The ML-engineering evaluation will run fixed competition subsets with identical token and time limits and report valid-submission rate, competition-specific score, refinement acceptance rate, execution failures, tokens, and latency. Each aggregate must be generated from machine-readable run artifacts, and no aggregate may be copied into the paper by hand.

\section{Discussion}
\label{sec:discussion}

\paragraph{Limitations.}
The list below states the known limits of this work.
\begin{itemize}[leftmargin=*,nosep,beginpenalty=10000]
  \item \textbf{Estimated coordinates in the default path.} The default scene graph does repeatable arithmetic over model-estimated two-dimensional coordinates. The arithmetic is deterministic, but its inputs are not measured, and deterministic arithmetic cannot correct a wrong geometric input.
  \item \textbf{Benchmark-fitted parser.} The strict parser was written against Q-Spatial Bench question text, with hand-written aliases, three unsupported rows, and a grounding-query override table. It is therefore not a held-out component.
  \item \textbf{Tuned thresholds.} The 0.30 instance-area ratio was chosen by inspecting development cases. Neither its effect on gap error nor the effect of stray reconstructed points on the gap has been evaluated.
  \item \textbf{External perception dependencies.} The metric path requires an externally installed SpatialClaw package and a reachable GPU perception service. Neither one is vendored, so the repository alone cannot run that path.
  \item \textbf{Unbundled data and controls.} The gated benchmark images and the frozen shuffled-image control artifact are not distributed with the repository.
  \item \textbf{No combined orchestrator.} Public source provides the four run modes and their validators, and it does not include the single-root orchestration that produced V37.
  \item \textbf{Inactive information-gain controller.} Expected-information-gain action selection is formulated here and present as a component, but no code in the repository calls it.
  \item \textbf{Uncalibrated and unbounded self-grade.} The self-grade that gates refinement can be any float, and a parse failure defaults to 0.5 and forces the refinement. It has not been shown to be calibrated and should not be read as a probability.
  \item \textbf{Optional detector.} Florence-2 runs only when its optional packages are installed, so a default install skips it.
  \item \textbf{Unvalidated FieldWorkArena adapter.} The adapter was never run against the benchmark, whose data remained inaccessible, so it must not be presented as evaluated compatibility.
  \item \textbf{Proxy validation score.} The ML path parses a validation score that the generated code prints. That score is a proxy supplied by the pipeline under test, and no outside check verifies it as a benchmark metric.
  \item \textbf{Incomplete executor isolation.} The execution controls are defense in depth, and they do not form a complete security sandbox.
  \item \textbf{Hand-designed templates and narrow leak coverage.} The strategy templates target common competition types and may not cover new tasks, and the leak audit covers only four leakage shapes.
  \item \textbf{No cancellation.} The A2A cancel operation is not supported once work is under way.
  \item \textbf{No native persistent loop and no answer-level verification.} The system has no native integration of the SpatialClaw persistent kernel. It also has no journal that records which evidence each final answer used, and no verifier that checks an answer against captured evidence.
  \item \textbf{No scientific comparison.} No protected score, paired arm comparison, uncertainty interval, resource-use result, or latency result has been produced. Accuracy, superiority, calibration, generalization, and production readiness are all unestablished.
\end{itemize}

\paragraph{Future work.}
We see five directions for future work.
\begin{itemize}[leftmargin=*,nosep]
  \item \textbf{Domain-specific fine-tuning.} Fine-tuning Florence-2 on industrial imagery may improve object detection for domain-specific objects such as safety equipment, pallet types, and industrial signage.
  \item \textbf{Multi-agent collaboration.} The A2A protocol allows architectures in which specialized sub-agents handle sub-tasks, such as visual analysis, spatial computation, and language generation.
  \item \textbf{Richer progress reporting.} The server already streams task status updates. Streaming the training logs of generated pipelines would give finer feedback during long runs.
  \item \textbf{Evidence integration.} Native SpatialClaw persistent-kernel integration, an answer-level evidence-use journal, and a verifier that checks answer claims against captured evidence all remain proposed.
  \item \textbf{Expanded benchmarks.} Extending the architecture to other agent benchmarks, such as SWE-bench \citep{jimenez2024swebench}, would test how general the design is.
\end{itemize}

\paragraph{Broader impact.}
The spatial scene graph approach could apply to industrial safety, where automated checks of clearance distances, equipment placement, and emergency exit access could help prevent workplace injuries. Automated spatial reasoning systems must be deployed with human oversight, because errors in safety-critical applications can have severe consequences.

\section{Conclusion}
\label{sec:conclusion}

We have presented Spatial Atlas, a research agent built on compute-grounded reasoning (CGR) and exposed through an A2A protocol server. Its implemented and proposed parts are listed below.
\begin{enumerate}[leftmargin=*,nosep]
  \item The \textbf{spatial scene-graph engine} makes extracted entities, coordinate assumptions, and computed relationships inspectable, and it does not claim that deterministic computation removes perception errors.
  \item The \textbf{confidence-gated refinement} lets one Fast-tier self-grade trigger at most one Strong-tier refinement. The entropy-guided information-gain selector is a proposed design. No code in the repository calls it, and this paper reports no ablation of its effect.
  \item The \textbf{fail-closed ML pipeline} combines strategy-aware code generation, explicit execution authorization, bounded repair attempts, and a separately gated dummy fallback.
  \item The \textbf{score-driven refinement loop} parses validation scores, requests a revision from the Strong tier, and keeps the revision only when its parsed score improves.
  \item The \textbf{leak audit and hint registry} prompts generated pipelines to check four common leakage patterns, and its one entry adds a targeted hint for a single competition.
  \item The \textbf{strict metric perception bridge} computes horizontal gaps from segmentation masks and a reconstructed point map, and it passes each gap to the model as a fact. It records provenance and evidence digests, and it fails closed when an evidence check fails.
  \item The \textbf{label-free run modes and validators} support a four-path comparison. They include a digest-bound, bijective, fixed-point-free shuffled-image control mapping and append-only journals that reject label-bearing fields.
\end{enumerate}

We also report one completed label-free operational run, V37, in which all four paths ran and the execution, warning-scan, restoration, cleanup, and terminal-seal gates passed. Each path produced eight prediction rows, for 32 rows in total, and protected labels and ground truth stayed sealed throughout. That run establishes operational integrity only. It is not a benchmark result, and it does not rank the paths against one another.

Compute-grounded reasoning has code compute the sub-problems that have deterministic solutions before the model writes its answer. This structure offers one way to make agent decisions easier to inspect. Whether that structure improves accuracy, reliability, latency, or resource use remains unanswered here. The next phase is the scientific one. It will unseal the protected labels under a frozen analysis plan, score the completed prediction journals, and report paired comparisons with uncertainty intervals. Only that step can turn the present operational evidence into a claim about performance.

The public code is at \url{https://github.com/arunshar/spatial-atlas-agent}. It includes the agent, the four run modes, and their validators, but it cannot re-execute V37 by itself (\cref{sec:repro}).

\appendix
\crefalias{section}{appendix}

\section{Reproducibility and Disclosure Boundary}
\label{sec:repro}

\paragraph{Code and presentation materials.}
The public code is available at \url{https://github.com/arunshar/spatial-atlas-agent}. The narrated demonstration is available at \url{https://www.youtube.com/watch?v=_0O3Mz0BUiw}. The presentation slides are available at \url{https://claude.ai/artifact/4AddsBUXqpmUCNYzkemxs2}. The video and slides explain the system, and the reproducible components are specified below.

\paragraph{Public entry points.}
The A2A server and task router are in \texttt{src/server.py} and \texttt{src/agent.py}. The spatial path is in \texttt{src/fieldwork/}. In that folder, \texttt{spatial.py} builds the scene graph, \texttt{perception.py} implements the strict metric bridge and the generic metric engine, \texttt{handler.py} selects the path, and \texttt{reasoner.py} implements the controller of \cref{alg:controller}. The self-grade and the unused information-gain selector are in \texttt{src/entropy/engine.py}. The ML path is in \texttt{src/mlebench/}, and the benchmark driver that exposes the run modes is \texttt{eval\_bench.py}.

\paragraph{Run-mode definitions.}
A path is selected by two options that are not freely combinable. The engine is either \texttt{scenegraph} or \texttt{metric}. The image mode is one of \texttt{correct}, \texttt{question-only}, or \texttt{shuffled}. The driver enforces four constraints. The \texttt{question-only} mode requires the \texttt{scenegraph} engine, since no image is supplied to measure. The \texttt{shuffled} mode requires the \texttt{metric} engine and an explicit control-mapping artifact. A control mapping is rejected for any image mode other than \texttt{shuffled}. The four paths in \cref{sec:evaluation} are therefore \texttt{scenegraph}+\texttt{correct}, \texttt{scenegraph}+\texttt{question-only}, \texttt{metric}+\texttt{correct}, and \texttt{metric}+\texttt{shuffled}, as in \cref{eq:arms}. The driver turns on strict mode whenever the \texttt{metric} engine is selected.

\paragraph{Control-mapping contract.}
In the public driver, a shuffled-image run is admitted only when the mapping holds exactly 84 pairs, is a bijection, contains no fixed point, and matches the recorded digest of every source and control image (\cref{eq:mapping}). Any drift, self-pairing, repeated source, or non-bijective mapping is rejected before the run starts.

\paragraph{Label-free journal schema.}
Pre-scoring rows are appended under the versioned label-free schema \texttt{label\_free\_prescore\_v1}. Each row stores the model's prediction text with run metadata and no ground-truth field. The writer rejects label-bearing keys anywhere in a payload. It also refuses to mix label-free and ordinary rows in one journal, and it refuses to resume a journal whose schema does not match the current run. These checks let a run complete and be inspected before any label is opened, and they check only what the run writes.

\paragraph{Test boundaries.}
The public test suite checks software behavior, and \cref{sec:tests} reports its totals. This paper does not describe what each test checks. Passing tests establish only the behavior they exercise. They are not benchmark evidence, and they do not promote proposed behavior into a result.

\paragraph{What is not reproducible from this repository.}
The metric path needs an externally installed SpatialClaw package and a reachable GPU perception service. Every V37 arm also needs a SpatialClaw checkout and the benchmark data, because the driver loads questions and images through the SpatialClaw benchmark factory. The strict arms also need a benchmark loader for the horizontal-gap questions, and neither public SpatialClaw nor this repository includes it. The gated benchmark images and the frozen control artifact are not bundled. The combined orchestration used for V37 ran in a separate private execution environment and is not included. So V37 cannot be re-executed from the public repository alone.

\paragraph{Disclosure boundary.}
This paper omits private execution identifiers, scheduler job identifiers, filesystem locations, credential material, service registry entries, raw service logs, private control digests, and protected labels or ground truth. Their absence is a disclosure decision. It does not remove evidence that any claim in this paper depends on.

\paragraph{Scope of reported numbers.}
The counts of 323 passed and 1 deselected come from a local run of the public CI test command, and the 92.09\% coverage figure covers \texttt{src/fieldwork/perception.py} only. The counts of 350 tests plus 63 subtests, and the V37 arm, row, retry, log, and finalizer-invocation counts, come from the private V37 record, which this repository does not include. The other numbers in the body are configured constants, code-level rules, and dependency versions read from the public source, plus the stated scale of a cited benchmark.

\paragraph{Revision note.}
This version removes the evaluation tables of the previous version, because no run artifact backs the values they reported. It also corrects the system description and the bibliography against the released code.

\end{document}

%% file: figures/fig_architecture.tex
\begin{figure}[t]
\centering
\definecolor{saInk}{HTML}{1F2A37}
\definecolor{saMain}{HTML}{2F5D8A}
\definecolor{saMainWash}{HTML}{EAF0F7}
\definecolor{saGray}{HTML}{6B7280}
\definecolor{saGrayWash}{HTML}{F3F4F6}
\definecolor{saOpt}{HTML}{9A6212}
\definecolor{saOptWash}{HTML}{FBF3E4}
\definecolor{saStop}{HTML}{A33A2B}
\definecolor{saStopWash}{HTML}{F9E9E6}
\begin{tikzpicture}[
    font=\footnotesize,
    execute at begin node={\hyphenpenalty=10000\relax},
    >={Stealth[length=1.7mm,width=1.3mm]},
    every node/.style={text=saInk},
    base/.style={rectangle, rounded corners=2pt, line width=0.6pt, align=center, inner sep=2.5pt},
    main/.style={base, draw=saMain, fill=saMainWash},
    opt/.style={base, draw=saOpt, fill=saOptWash},
    stop/.style={base, draw=saStop, fill=saStopWash},
    neutral/.style={base, draw=saGray, fill=saGrayWash},
    colA/.style={main, text width=4.3cm},
    colB/.style={text width=3.85cm},
    flow/.style={->, line width=0.6pt, draw=saInk},
    optflow/.style={->, line width=0.6pt, draw=saOpt, dashed},
    stopflow/.style={->, line width=0.6pt, draw=saStop},
    shareflow/.style={->, line width=0.6pt, draw=saGray, dashed},
    lab/.style={font=\scriptsize, inner sep=1pt},
  ]

  \node[neutral, minimum width=13.7cm, minimum height=0.55cm] (a2a) at (6.85,0)
    {\textbf{A2A server} built on the official \texttt{a2a-sdk}, with \textbf{rule-based dispatch}: four fixed rules and no model call};

  \node[colA, anchor=north] (parse) at (2.4,-1.75)
    {\textbf{File parsing into evidence}\\[1pt]{\scriptsize The Vision tier describes images and video frames, and pypdf reads PDF text.}};
  \node[colA, below=0.45cm of parse] (scene)
    {\textbf{Scene extraction (Strong tier)}\\[1pt]{\scriptsize Positions are Strong-tier estimates from text. Model-supplied distances are kept.}};
  \node[colA, below=0.45cm of scene] (facts)
    {\textbf{Computed facts (code)}\\[1pt]{\scriptsize Code fills only the missing distances and checks the extracted safety rules.}};
  \node[colA, below=0.45cm of facts] (answer)
    {\textbf{Answer (Strong tier)}\\[1pt]{\scriptsize It reads the question, the file descriptions, and the fact sheet.}};
  \node[colA, below=0.5cm of answer] (grade)
    {\textbf{Self-grade (Fast tier)}\\[1pt]{\scriptsize It returns $\sigma$, or 0.5 when the reply cannot be parsed.}};
  \node[colA, below=0.45cm of grade] (fmt)
    {\textbf{Output formatter}};

  \draw[flow] (a2a.south -| parse.north) -- (parse.north);
  \node[lab, anchor=west] at (2.5,-0.6) {question format, or default};
  \draw[flow] (parse) -- (scene) node[lab, midway, right, xshift=1.5pt] {file descriptions};
  \draw[flow] (scene) -- (facts) node[lab, midway, right, xshift=1.5pt] {typed scene graph};
  \draw[flow] (facts) -- (answer) node[lab, midway, right, xshift=1.5pt] {fact sheet};
  \draw[flow] (answer) -- (grade);
  \draw[flow] (grade) -- (fmt) node[lab, midway, right, xshift=1.5pt] {$\sigma \ge 0.6$};

  \path (fmt.south) ++(0,-0.15) coordinate (fwbot);
  \draw[saMain, line width=0.6pt, rounded corners=3pt] (0,-0.92) rectangle (fwbot -| 9.6,0);
  \node[anchor=north east, font=\footnotesize\bfseries, text=saMain] at (9.5,-0.94) {Spatial handler};

  \node[neutral, dashed, colB, anchor=north] (tools) at (7.3,-2.85)
    {\textbf{SpatialClaw tools}\\[1pt]{\scriptsize These external tools supply SAM~3 masks and a Depth Anything~3 point map.}};
  \node[opt, colB, below=0.3cm of tools] (gap)
    {\textbf{Horizontal surface gap}\\[1pt]{\scriptsize The checks of \Cref*{fig:bridge} fail closed. A valid gap enters the fact sheet in place of the scene graph.}};
  \node[stop, colB, below=0.3cm of gap] (exits)
    {{\scriptsize A failed evidence check returns \texttt{measurement unavailable} and skips the reasoner.\\[1pt] A parse, manifest, or service failure raises an error.}};
  \path (exits.south) ++(0,-0.15) coordinate (brbot);
  \draw[saOpt, line width=0.6pt, dashed, rounded corners=3pt] (5.15,-1.7) rectangle (brbot -| 9.45,0);
  \node[anchor=north west, align=left, inner sep=2pt] at (5.2,-1.72)
    {\textbf{\color{saOpt}Strict metric bridge}\\[-1pt]{\scriptsize benchmark driver only,}\\[-1pt]{\scriptsize never reached through A2A}};
  \draw[optflow] (2.4,-1.42) -- (8.95,-1.42) -- (8.95,-1.7);
  \node[lab, anchor=south west, text=saOpt] at (2.5,-1.4) {driver only: image and raw question};
  \draw[flow] (tools) -- (gap);
  \draw[stopflow] (gap) -- (exits);
  \draw[flow] (gap.west) -| (4.9,0 |- answer.east) -- (answer.east);

  \node[main, text width=3.3cm] (refine) at (grade -| 7.55,0)
    {\textbf{One refinement (Strong tier)}\\[1pt]{\scriptsize It runs at most once per task.}};
  \draw[flow] (grade) -- (refine) node[lab, midway, above] {$\sigma < 0.6$};
  \draw[flow] (refine.south) |- (fmt.east);

  \node[neutral, text width=3.7cm, anchor=north] (mle) at (11.8,-0.92)
    {\textbf{ML-engineering handler}\\[2pt]
     {\scriptsize The Standard tier analyzes the competition and names a strategy template.\\[2pt]
      The Strong tier writes the pipeline script, repairs it after errors, and refines it.\\[2pt]
      Generated-code execution is off by default. It needs an opt-in flag, an operator flag that attests an isolated and trusted worker, and authenticated startup.}};
  \draw[flow] (a2a.south -| mle.north) -- (mle.north) node[lab, midway, right, xshift=2pt, align=left] {archive, or\\ML keywords};

  \node[neutral, minimum width=13.7cm, minimum height=0.5cm, anchor=north] (shared) at ([yshift=-0.4cm]fwbot -| 6.85,0)
    {\textbf{Shared by both handlers}: {\scriptsize LiteLLM client, four model tiers (Fast, Standard, Strong, Vision), and usage tracking}};
  \draw[shareflow] (fwbot -| 4.8,0) -- (shared.north -| 4.8,0);
  \draw[shareflow] (mle.south) -- (mle.south |- shared.north);
\end{tikzpicture}
\caption{Spatial Atlas places two handlers behind one A2A server. The server passes each task to one handler through four fixed dispatch rules, and no model call takes part in dispatch. On the default spatial path, the Strong tier extracts a typed scene graph from the file descriptions, so entity positions are Strong-tier estimates from text. Code computes only the distances that the model did not supply, and the Strong tier answers from the resulting fact sheet. A Fast-tier self-grade below 0.6 triggers one Strong-tier refinement. The strict metric bridge runs only under the frozen benchmark driver, which supplies the raw question and the protocol identifier. It calls SpatialClaw's tools~\citep{cho2026spatialclaw} for SAM~3 masks~\citep{carion2025sam3} and Depth Anything~3 point maps~\citep{lin2025da3}. The bridge fails closed and never falls back to the scene graph. The information-gain selector and the tier router exist in code, but no code calls them, so the figure omits them.}
\label{fig:architecture}
\end{figure}

%% file: figures/fig_bridge.tex
\begin{figure}[t]
\centering
\definecolor{saInk}{HTML}{1F2A37}
\definecolor{saMain}{HTML}{2F5D8A}
\definecolor{saMainWash}{HTML}{EAF0F7}
\definecolor{saGray}{HTML}{6B7280}
\definecolor{saGrayWash}{HTML}{F3F4F6}
\definecolor{saOpt}{HTML}{9A6212}
\definecolor{saOptWash}{HTML}{FBF3E4}
\definecolor{saStop}{HTML}{A33A2B}
\definecolor{saStopWash}{HTML}{F9E9E6}
\begin{tikzpicture}[
    font=\footnotesize,
    execute at begin node={\hyphenpenalty=10000\relax},
    >={Stealth[length=1.7mm,width=1.3mm]},
    every node/.style={text=saInk},
    gstep/.style={rectangle, rounded corners=2pt, line width=0.6pt, align=center, inner sep=2.5pt,
      draw=saOpt, fill=saOptWash, text width=2.3cm, minimum height=1.8cm},
    check/.style={gstep, line width=1.0pt},
    flow/.style={->, line width=0.6pt, draw=saInk},
    exit/.style={->, line width=0.6pt, draw=saStop},
    pairexit/.style={->, line width=0.6pt, draw=saStop, dashed},
    rail/.style={line width=0.6pt, draw=saStop},
    lab/.style={font=\scriptsize, inner sep=1pt},
    exitlab/.style={font=\scriptsize, inner sep=1pt, text=saStop},
  ]

  \node[check] (in) at (1.24,0)
    {\textbf{Inputs}\\[1pt]{\scriptsize SAM~3 gives masks, and Depth Anything~3 gives a point map.}};
  \node[check] (erode) at (3.97,0)
    {\textbf{Erosion}\\[1pt]{\scriptsize Each mask is resized and then eroded once with a disk of radius $r_e$.}};
  \node[check] (overlap) at (6.70,0)
    {\textbf{Overlap check}\\[1pt]{\scriptsize A pair with $\rho \ge 0.05$ fails. Otherwise the shared pixels are removed.}};
  \node[gstep] (conf) at (9.43,0)
    {\textbf{Confidence}\\[1pt]{\scriptsize Finite points with confidence above 0.3 stay.}};
  \node[gstep] (voxel) at (12.16,0)
    {\textbf{5 mm voxels}\\[1pt]{\scriptsize Each $(x, z)$ voxel keeps its top-confidence point.}};

  \node[check] (count) at (1.24,-2.8)
    {\textbf{Voxel count}\\[1pt]{\scriptsize Each object needs at least 32 voxels.}};
  \node[gstep] (nn) at (3.97,-2.8)
    {\textbf{Directed $\operatorname{pct}_5$}\\[1pt]{\scriptsize Nearest-neighbor distances run $A \to B$ and $B \to A$.}};
  \node[gstep] (gap) at (6.70,-2.8)
    {\textbf{Gap}\\[1pt]{\scriptsize $g$ is the smaller directed value.}};
  \node[check] (valid) at (9.43,-2.8)
    {\textbf{Validity}\\[1pt]{\scriptsize $g$ must be finite, with $0 < g \le 20$ m.}};
  \node[gstep] (select) at (12.16,-2.8)
    {\textbf{Pair selection}\\[1pt]{\scriptsize The smallest valid $g$ wins, and ties go to the lowest mask indices.}};

  \draw[flow] (in) -- (erode);
  \draw[flow] (erode) -- (overlap);
  \draw[flow] (overlap) -- (conf);
  \draw[flow] (conf) -- (voxel);
  \draw[flow] (voxel.south) -- (12.16,-1.4) -- (1.24,-1.4) -- (count.north);
  \draw[flow] (count) -- (nn);
  \draw[flow] (nn) -- (gap);
  \draw[flow] (gap) -- (valid);
  \draw[flow] (valid) -- (select);

  \node[rectangle, rounded corners=2pt, line width=1.0pt, draw=saStop, fill=saStopWash, align=center,
        inner sep=3pt, text width=3.4cm] (unavail) at (2.0,-5.8)
    {\textbf{Fail closed}\\[1pt]\texttt{QSPATIAL\_UNAVAILABLE}\\{\scriptsize The answer is the literal \mbox{``measurement unavailable''}.}};
  \node[rectangle, rounded corners=2pt, line width=1.0pt, draw=saMain, fill=saMainWash, align=center,
        inner sep=3pt, text width=2.6cm] (answer) at (12.16,-5.8)
    {\textbf{Gap fact}\\[1pt]{\scriptsize $g$ enters the fact sheet to six decimals. The Strong tier writes the answer.}};
  \draw[flow] (select) -- (answer);
  \node[lab, align=center, text width=5.6cm] at (7.0,-5.85)
    {Dashed exits drop only the current pair. The bridge fails closed when no pair stays valid.};

  \draw[rail] (in.north) -- (1.24,1.5) -- (-0.3,1.5) -- (-0.3,-5.8);
  \draw[exit] (-0.3,-5.8) -- (unavail.west);
  \node[exitlab, anchor=west] at ([xshift=1pt,yshift=3.6mm]in.north) {too few masks};
  \draw[exit] ([xshift=-7.6mm]select.south) |- (2.0,-4.5) -- (unavail.north);
  \node[exitlab, anchor=south] at (5.3,-4.5) {no valid pair};

  \foreach \n/\lbl in {erode/empty after erosion, overlap/{$\rho \ge 0.05$}} {
    \draw[pairexit] (\n.north) -- ++(0,0.5);
    \node[exitlab, anchor=west] at ([xshift=1pt,yshift=2.8mm]\n.north) {\lbl};
  }
  \draw[pairexit] (count.south) -- ++(0,-0.5);
  \node[exitlab, anchor=west] at ([xshift=1pt,yshift=-2.8mm]count.south) {fewer than 32 voxels};
  \draw[pairexit] (valid.south) -- ++(0,-0.5);
  \node[exitlab, anchor=east] at ([xshift=-1pt,yshift=-2.8mm]valid.south) {non-finite, or outside $(0, 20]$ m};
\end{tikzpicture}
\caption{The strict metric bridge computes a horizontal gap from object masks and a point map. Each candidate pair of object masks passes through the steps from left to right, first along the top row and then along the bottom row. Here $\operatorname{pct}_5$ is the fifth percentile with linear interpolation, as in \cref{eq:gapfull}. The erosion radius $r_e$ is the ceiling of the larger resolution ratio between a mask and the point map. Each mask is first resized to the point-map grid by nearest-neighbor sampling. The overlap coefficient is $\rho = \lvert E_a \cap E_b \rvert / \min(\lvert E_a \rvert, \lvert E_b \rvert)$ for the eroded masks $E_a$ and $E_b$. The bridge returns \texttt{QSPATIAL\_UNAVAILABLE} at once when the parser marks the question unsupported or when segmentation returns too few masks. A parse or manifest failure raises an error before the steps in this figure begin. Every dashed exit drops only the current pair, and the bridge returns \texttt{QSPATIAL\_UNAVAILABLE} when no pair stays valid. The confidence filter has no exit of its own, because a mask that loses its points fails the voxel count. A segmentation or reconstruction service failure raises an error, and the bridge never falls back to the scene graph. A valid gap enters the fact sheet, and the Strong tier writes the final answer from it.}
\label{fig:bridge}
\end{figure}

%% file: figures/fig_arms.tex
\begin{figure}[t]
\centering
\definecolor{saInk}{HTML}{1F2A37}
\definecolor{saMain}{HTML}{2F5D8A}
\definecolor{saMainWash}{HTML}{EAF0F7}
\definecolor{saGray}{HTML}{6B7280}
\definecolor{saGrayWash}{HTML}{F3F4F6}
\definecolor{saOpt}{HTML}{9A6212}
\definecolor{saOptWash}{HTML}{FBF3E4}
\definecolor{saStop}{HTML}{A33A2B}
\definecolor{saStopWash}{HTML}{F9E9E6}
\begin{tikzpicture}[
    font=\footnotesize,
    execute at begin node={\hyphenpenalty=10000\relax},
    >={Stealth[length=1.7mm,width=1.3mm]},
    every node/.style={text=saInk},
    base/.style={rectangle, rounded corners=2pt, line width=0.6pt, align=center, inner sep=2.5pt},
    main/.style={base, draw=saMain, fill=saMainWash},
    opt/.style={base, draw=saOpt, fill=saOptWash},
    stop/.style={base, draw=saStop, fill=saStopWash},
    neutral/.style={base, draw=saGray, fill=saGrayWash},
    arm/.style={text width=2.27cm, minimum height=1.6cm, anchor=north},
    flow/.style={->, line width=0.6pt, draw=saInk},
    lab/.style={font=\scriptsize, inner sep=1pt},
  ]

  \node[main, arm] (a1) at (1.225,0)
    {\textbf{Scene-graph\\path}\\[1pt]{\scriptsize engine: scene graph\\image: correct}};
  \node[main, arm] (a2) at (3.925,0)
    {\textbf{Question-only\\baseline}\\[1pt]{\scriptsize engine: scene graph\\image: none}};
  \node[opt, arm] (a3) at (6.625,0)
    {\textbf{Correct-image\\metric path}\\[1pt]{\scriptsize engine: strict metric\\image: correct}};
  \node[opt, arm] (a4) at (9.325,0)
    {\textbf{Shuffled-image\\metric control}\\[1pt]{\scriptsize engine: strict metric\\image: control $\phi(s)$}};

  \foreach \x/\col/\wash in {1.225/saMain/saMainWash, 3.925/saMain/saMainWash, 6.625/saOpt/saOptWash, 9.325/saOpt/saOptWash} {
    \foreach \k in {0,...,7} {
      \draw[draw=\col, fill=\wash, line width=0.5pt] ({\x-1.055+\k*0.27},-2.05) rectangle ++(0.22,-0.34);
    }
    \draw[flow] (\x,-1.6) -- (\x,-2.0);
    \node[lab] at (\x,-2.63) {8 label-free rows};
    \draw[flow] (\x,-2.78) -- (\x,-3.18);
  }

  \node[neutral, minimum width=13.75cm, minimum height=0.6cm] (journal) at (6.875,-3.48)
    {\textbf{Label-free prescore journals}: {\scriptsize 32 label-free prescore prediction rows, with zero retries}};
  \node[neutral, text width=3.9cm, minimum height=1.45cm, anchor=north west] (final) at (0,-4.2)
    {\textbf{Terminal finalizer}\\[1pt]{\scriptsize It was invoked once and wrote the smoke seal and the terminal result.}};
  \node[neutral, text width=2.5cm, minimum height=1.45cm, anchor=north west] (seal) at (4.48,-4.2)
    {\textbf{Seal and result}\\[1pt]{\scriptsize They form the closed final record of the run.}};
  \draw[flow] (journal.south -| final.north) -- (final.north);
  \draw[flow] (final) -- (seal);

  \node[stop, text width=2.7cm, minimum height=1.45cm, anchor=north west] (labels) at (7.71,-4.2)
    {\textbf{Labels sealed}\\[1pt]{\scriptsize Protected labels and ground truth remained sealed.}};
  \node[neutral, dashed, text width=2.5cm, minimum height=1.45cm, anchor=north east] (score) at (13.77,-4.2)
    {\textbf{No scoring}\\[1pt]{\scriptsize No score or paired comparison was computed.}};
  \draw[line width=0.6pt, draw=saGray, dashed] (labels) -- (score)
    node[midway, fill=white, inner sep=0.5pt, text=saStop, font=\footnotesize\bfseries] {$\times$};

  \draw[saGray, line width=0.6pt, rounded corners=3pt] (10.85,0) rectangle (13.77,-2.97);
  \node[anchor=north, align=center, inner sep=2pt] at (12.31,-0.05) {\textbf{Public driver}\\\textbf{contract $\phi$}};
  \foreach \i in {0,...,5} {
    \coordinate (m\i) at ({12.31+0.36*cos(90-60*\i)},{-1.24+0.36*sin(90-60*\i)});
  }
  \foreach \i/\j in {0/1,1/2,2/3,3/4,4/5} {
    \draw[-{Stealth[length=1.1mm,width=1.0mm]}, line width=0.5pt, draw=saOpt, shorten <=1.0mm, shorten >=1.0mm] (m\i) to[bend left=25] (m\j);
  }
  \draw[-{Stealth[length=1.1mm,width=1.0mm]}, line width=0.5pt, draw=saOpt, dotted, shorten <=1.0mm, shorten >=1.0mm] (m5) to[bend left=25] (m0);
  \foreach \i in {0,...,5} {
    \fill[saInk] (m\i) circle (0.045);
  }
  \node[lab, anchor=south west] at ([xshift=1pt,yshift=0.5pt]m1) {$s$};
  \node[lab, anchor=north west] at ([xshift=1pt,yshift=-0.5pt]m2) {$\phi(s)$};
  \node[lab, align=center, text width=2.75cm, anchor=north] at (12.31,-1.72)
    {The public code requires 84 pairs, a bijection, no fixed point, and checked image digests.};
  \draw[flow] (10.85,-0.8) -- (a4.east |- 0,-0.8);
\end{tikzpicture}
\caption{Four operational paths ran together in the V37 run. The paths were the scene-graph path, question-only baseline, correct-image metric path, and schema-matched shuffled-image metric control. Each path produced eight label-free prescore prediction rows, so the run wrote 32 rows with zero retries. The execution, warning-scan, restoration, cleanup, and terminal-seal gates passed. In the public driver, the shuffled-image mode swaps each attached image through a control mapping $\phi$. That mapping must hold 84 pairs, be a bijection, and have no fixed point, and the loader checks the digest of every source and control image. The terminal finalizer ran once and wrote the smoke seal and the terminal result. Protected labels and ground truth remained sealed, and no score or paired comparison was computed, so the run establishes operational integrity only. The run is private, and the public repository cannot reproduce it.}
\label{fig:arms}
\end{figure}